\documentclass[10pt,twocolumn,letterpaper]{article}

\usepackage{iccv}
\usepackage{times}
\usepackage{epsfig}
\usepackage{graphicx}
\usepackage{amsmath}
\usepackage{amssymb}
\usepackage{multicol}
\usepackage{array}
\usepackage{float}
\usepackage{enumitem}

\usepackage{booktabs}
\usepackage{multirow}
\usepackage{tabu}
\usepackage{makecell}
\usepackage[titletoc]{appendix}

\usepackage[pagebackref=true,breaklinks=true,letterpaper=true,colorlinks,bookmarks=false]{hyperref}

\iccvfinalcopy 

\def\iccvPaperID{0001} 

\ificcvfinal\fi
\begin{document}

\title{MixMatch Domain Adaptaion: Prize-winning solution for both tracks of \\ VisDA 2019 challenge}

\author{Danila Rukhovich\\
{\tt\small d.rukhovich@samsung.com}
\and
Danil Galeev\\
{\tt\small d.galeev@samsung.com}
}

\maketitle
\thispagestyle{empty}

\begin{abstract}
We present a domain adaptation (DA) system that can be used in multi-source and semi-supervised settings. Using the proposed method we achieved \textbf{2nd} place on multi-source track and \textbf{3rd} place on semi-supervised track of the VisDA 2019 challenge \footnote{\url{http://ai.bu.edu/visda-2019/}}.
The source code of out method is available publicly \footnote{\url{https://github.com/filaPro/visda2019}}.
\end{abstract}

\section{Introduction}

Unsupervised domain adaptation aims to generalize a model learned from a source domain with rich annotated data to a new target domain without any labeled data. To speed up research progress in this area the DomainNet \cite{domainnet} dataset was released. This dataset contains around 0.6 million images of 345 classes and 6 domains: $real$, $infograph$, $quickdraw$, $sketch$, $clipart$ and $painting$. The VisDA 2019 challenge aims to test new domain adaptation algorithms on this dataset. For multi-source track the task is to train model on image from 4 annotated domains and 2 unlabeled domains to maximize accuracy on these 2 target domains. For semi-supervised track a very few (3 per class) labeled images from 2 target domains are available for training and only $real$ is used as a source domain.

\section{Proposed method}

\begin{figure*}[h]
 \begin{tabular}{c}
    \includegraphics[width=\linewidth]{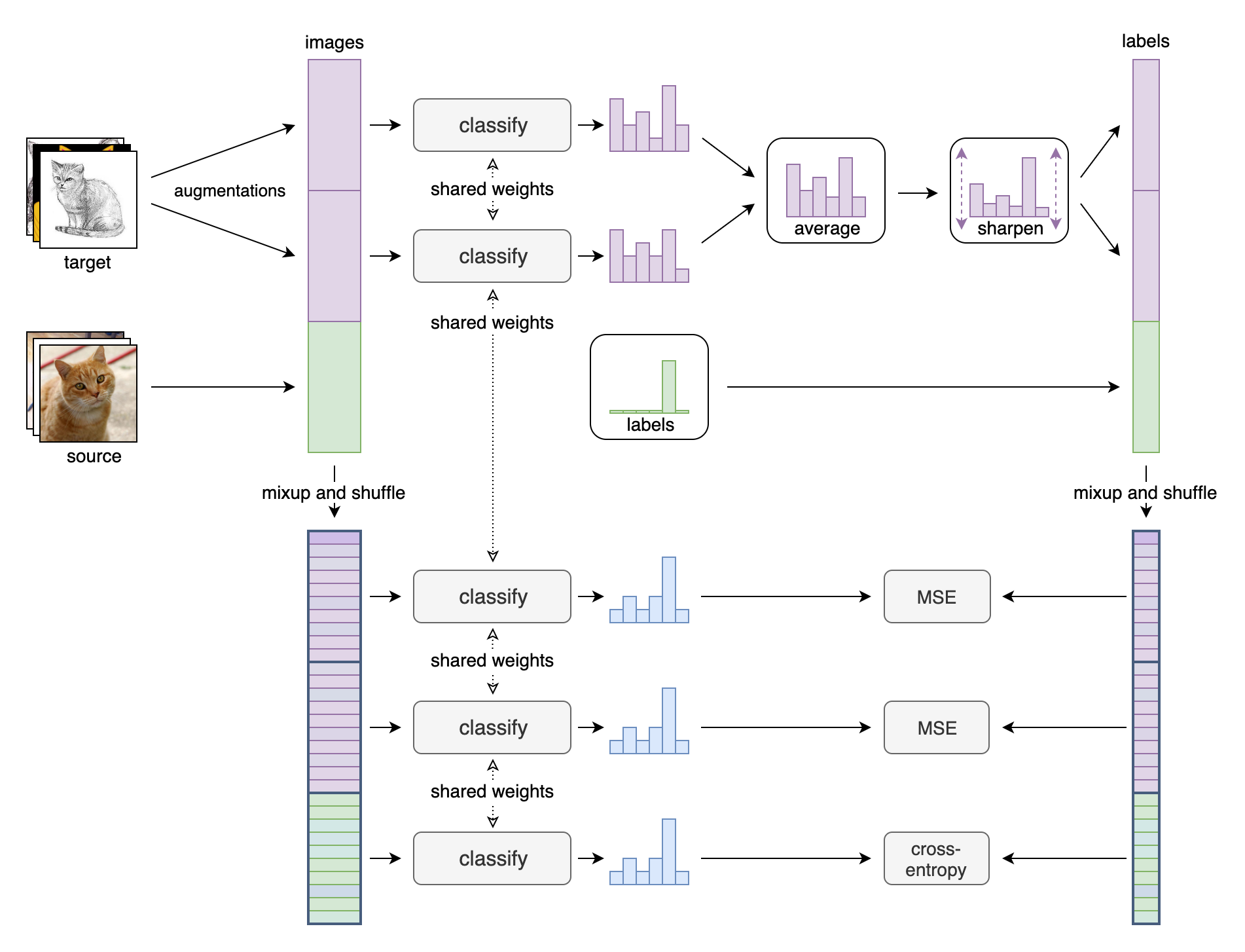} \\
  \end{tabular}
  \vspace{5pt}
  \caption{Scheme of the proposed method.}
  \label{fig:scheme}
\end{figure*}

In short, our method can be characterized as MixMatch with EfficientNet backbone. In this section we present a brief description of these 2 architectures and our contribution to their application to the contest tasks. General scheme of our approach is shown in Figure \ref{fig:scheme}.

\subsection{MixMatch}

MixMatch \cite{mixmatch} is a state-of-the-art method in semi-supervised learning. Its main idea is the combining of currently dominating approaches in this field like multiple augmentations, label guessing, label distribution sharpening and MixUp \cite{mixup}. The authors conducted experiments showing the effect of each of these components on the final classification accuracy. Borrowing this insights we show how this method can be applied to domain adaptation tasks.

\subsection{EfficientNet}

Although MixMatch and M3SDA \cite{domainnet} use ResNet architectures (Wide Resnet-28 and Resnet-101) as a backbone for their methods, we decided to use state-of-the-art ImageNet classification architecture EfficientNet \cite{efficientnet}. This model is a result of neural architecture search with carefully balancing of network depth, width and resolution. It is also shown that better resulting models from image classification and transfer learning have even less number of parameters. One more advantage of using this network is the set of 8 pretrained models: EfficientNet-b0, ..., EfficientNet-b7 with increasing number of parameters and overall accuracy. In our experiments first five models from this list didn't generalized well, and the last one EfficientNet-b7 was too heavy,
which necessitated a reduction in the size of the batches and resulted a decrease in accuracy.

\subsection{Overview}

As the MixMatch is not a domain adaptation method, we propose a new scheme for constructing mini-batches during training. Originally, batch contains $n$ labeled image, $n$ unlabeled images and $n$ same unlabeled images with different random augmentation. For our approach unlabeled part transforms to target domain with no changes. For multi-source domain adaptation we construct the labeled part of batch from $\frac{n}{k}$ images from each of $k$ source domains. For semi-supervised domain adaptation the labeled part of batch consists of $\frac{4n}{5}$ source images and $\frac{n}{5}$ images from labeled part of target domain.

We also propose a training process modification to improve its stability. During one MixMatch training step the backbone model is called 5 times: 2 times for target domain batches and 3 times for mixed up batches. The distributions in these 3 batches differ significantly, because one is dominated by source images and the other two by target images. This, in turn, causes instability when updating the batch normalization layer statistics. We form each batch of $\frac{1}{3}$ source and $\frac{2}{3}$ target images to approximate the statistics between the batches entering the network.

In addition to experimenting with the MixMatch model, we also tested our baseline. This approach only trains EfficientNet on all labeled data. Comparison of this 2 models is presented in section \ref{validation}.

We use the same data augmentation strategy during training and testing time:
\begin{enumerate}[nosep]
    \item resize to $256 \times 256$ pixels,
    \item random horizontal flip,
    \item random crop of size $224 \times 224$ pixels.
\end{enumerate}

\subsection{Model ensembling}

It is widely known that the ensemble averaging of neural networks trained independently leads to the improvement of test accuracy. In this work we trained models with different EfficientNet backbones and different weight of loss balancing (cross-entropy and mean squared). We used equal average of predictions from these models to make final prediction. In our experiments this technique gave an increase in accuracy for both tasks of the challenge.

\section{Experiments} \label{experiments}

\subsection{Training details}

We implemented our MixMatch-based and baseline models in Tensorflow 2.0 from scratch. We used open source EfficientNet pre-trained on ImageNet models \footnote{\url{https://github.com/qubvel/efficientnet}}. For all experiments without ensembling we used same hyper-parameters. We trained the network with Adam optimizer with $0.0001$ learning rate and batch size of $15$ for $100$ epochs (epoch is $1000$ batches; $10$ epochs is enough for baseline models). MixMatch parameters (except of loss weight) are set to their default values from the original paper, in particular, beta distribution parameter of $0.75$ and label sharpening power of $0.5$. Loss weight parameter is a multiplier for mean squared error, we set its default value to $333.0$. For all experiments we used $7$ test time augmentations.

We trained our models on 8 Nvidia Tesla P40 GPUs with 24 Gb memory each, which allows us to use $15\times3\times8$ images per one optimizer update step. We noticed that smaller batch size leads to decreasing of target accuracy.

\subsection{Validation} \label{validation}

During the validation phase of the competition $sketch$ was the target domain, for multi-source (ms) track $real$, $quickdraw$ and $infograph$ were the source domains and for semi-supervised (ss) track only $real$ was the source domain.

To demonstrate the benefits of domain adaptation, we compare the MixMatch model with baseline (not using DA). As can be seen from Table \ref{tab:validation}, the growth of target accuracy on both tracks is about $10\%$. For semi-supervised track we also show the benefit of using the labeled part of target domain even for baseline model. Here we use MixMatch model with EfficientNet-b5 backbone.

\begin{table}[h]
\begin{center}
\begin{tabular}{|c|c|c|}
\hline
track & model & accuracy \\ \hline
\hline
ms & baseline & $0.519$ \\ \hline
ms & MixMatch & $\boldsymbol{0.619}$ \\ \hline \hline
ss & baseline w/o labeled target & $0.457$ \\ \hline
ss & baseline & $0.520$ \\ \hline
ss & MixMatch & $\boldsymbol{0.639}$ \\ \hline
\end{tabular}
\end{center}
\caption{Evaluation of different models for validation phase.}
\label{tab:validation}
\end{table}

\subsection{Testing} \label{testing}

During the testing phase of the competition $clipart$ and $painting$ were used as the target domain, for multi-source track $real$, $quickdraw$, $infograph$ and $sketch$ were the source domains and for semi-supervised track only $real$ was the source domain. We trained same models for 2 target domains and then concatenated predictions for final submission.

Our results are shown in table \ref{tab:testing}. For both multi-source and semi-supervised tracks we trained 3 models with 2 different backbones and loss weights. The accuracy of all models is almost equal, and model ensembling gives $1\%$ profit.

\begin{table}[h]
\begin{center}
\begin{tabular}{|c|c|c|c|}
\hline
track & EfficientNet & loss weight & accuracy \\ \hline
\hline
ms & b5 & $333.0$ & $0.699$ \\ \hline
ms & b5 & $1000.0$ & $0.689$ \\ \hline
ms & b6 & $333.0$ & $0.695$ \\ \hline 
ms & \multicolumn{2}{c|}{ensemble} & {$\boldsymbol{0.716}$} \\ \hline \hline
ss & b5 & $333.0$ & $0.704$ \\ \hline
ss & b5 & $1000.0$ & $0.695$ \\ \hline
ss & b6 & $333.0$ & $0.703$ \\ \hline
ss & \multicolumn{2}{c|}{ensemble} & $\boldsymbol{0.713}$ \\ \hline
\end{tabular}
\end{center}
\caption{Evaluation of MixMatch model with different parameters for test phase.}
\label{tab:testing}
\end{table}

With these results, we achieved the prize-winning places of the VisDA 2019 challenge. Top 3 results for both tracks are shown in Table \ref{tab:leaderboard} (our team name is $denemmy$).

\begin{table}[h]
\begin{center}
\begin{tabular}{|c|c|c|}
\hline
track & team & accuracy \\ \hline
\hline
ms & Yingwei.Pan & $\boldsymbol{0.760}$ \\ \hline
ms & \textbf{denemmy} & $0.716$ \\ \hline
ms & numpee & $0.696$ \\ \hline \hline
ss & lunit & $\boldsymbol{0.720}$ \\ \hline
ss & Yingwei.Pan & $0.714$ \\ \hline
ss & \textbf{denemmy} & $0.713$ \\ \hline
\end{tabular}
\end{center}
\caption{Top 3 places on final leaderboard.}
\label{tab:leaderboard}
\end{table}

{\small
\bibliographystyle{ieee}
\bibliography{main}
}

\end{document}